\def\year{2022}\relax
\documentclass[letterpaper]{article} 
\usepackage{aaai22}  
\usepackage{times}  
\usepackage{helvet}  
\usepackage{courier}  
\usepackage[hyphens]{url}  
\usepackage{graphicx} 
\usepackage{amssymb}
\def\UrlFont{\rm}  
\usepackage{natbib}  
\usepackage{caption} 
\DeclareCaptionStyle{ruled}{labelfont=normalfont,labelsep=colon,strut=off} 
\usepackage{algorithm}
\usepackage{algorithmic}
\def\etal{\textit{et al}}
\usepackage{newfloat}
\usepackage{listings}
\floatstyle{ruled}
\newfloat{listing}{tb}{lst}{}
\floatname{listing}{Listing}

\title{AAAI Press Formatting Instructions \\for Authors Using \LaTeX{} --- A Guide}

\title{Controllable Face Manipulation and UV Map Generation by Self-supervised Learning}

\author {
    Yuanming Li \textsuperscript{\rm 1},
    Jeong-gi Kwak \textsuperscript{\rm 1}, 
    David Han\textsuperscript{\rm2}, 
    Hanseok Ko \textsuperscript{\rm 1}
}
\affiliations {
    \textsuperscript{\rm 1} Korea Univerisity\\
    \textsuperscript{\rm 2} Drexel University\\
    lym7499500@korea.ac.kr, 
    kjk8557@korea.ac.kr, 
    dkh42@drexel.edu,
    hsko@korea.ac.kr
}

\usepackage{bibentry}

\begin{document}

\maketitle

\begin{abstract}

Although manipulating facial attributes by Generative Adversarial Networks (GANs) has been remarkably successful recently, there are still some challenges in explicit control of features such as pose, expression, lighting, etc. Recent methods achieve explicit control over 2D images by combining 2D generative model and 3DMM. However, due to the lack of realism and clarity in texture reconstruction by 3DMM, there is a domain gap between the synthetic image and the rendered image of 3DMM. Since rendered 3DMM images contain facial region only without the background, directly computing the loss between these two domains is not ideal and the resultant trained model will be biased. In this study, we propose to explicitly edit the latent space of the pretrained StyleGAN by controlling the parameters of the 3DMM. To address the domain gap problem, we propose a noval network called ‘Map and edit’ and a simple but effective attribute editing method to avoid direct loss computation between rendered and synthesized images. Furthermore, since our model can accurately generate multi-view face images while the identity remains unchanged. As a by-product, combined with visibility masks, our proposed model can also generate texture-rich and high-resolution UV facial textures. Our model relies on pretrained StyleGAN, and the proposed model is trained in a self-supervised manner without any manual annotations or datasets.



\end{abstract}

\section{Introduction}
Modifying certain properties of a given portrait image, also known as face image processing, has many applications including photo retouching, UV map completion, virtual reality, etc. With rapid development of Generative Adversarial Networks (GAN) \cite{goodfellow2014generative}, State-of-the-Art (SOTA) models, such as StyleGAN, can generate realistic photo portraits \cite{karras2019style, karras2020analyzing}. 

High-fidelity image generation enabled by the latent representation of StyleGAN has attracted considerable attention as it helps facilitate controllable manipulation. Several recent studies \cite{shen2020interpreting, yang2021semantic} have shown that linear operations in the latent space of StyleGAN can control specific target attributes. In its latent space, the early and late layers control the image layout and final rendering, respectively. To enable more refined real-world image editing by StyleGAN, GAN inversion techniques \cite{tewari2020pie, richardson2021encoding,tov2021designing,wang2022high, lee2020uncertainty} have recently been explored, which aims to project images into the latent space of a pretrained GAN generator. 

Recently, many studies \cite{deng2018uv, gecer2019ganfit, tewari2020stylerig,zhou2020rotate, deng2020disentangled, gecer2021ostec, lin2021meingame, medin2022most} have been conducted to combine synthesis capabilities of GAN with the physics based 3D Morphable Model (3DMM) \cite{blanz1999morphable, bfm09}. This allows applying explicit 3D models as a guiding tool for editing face images (e.g. face rotation, expression and illumination) and completing realistic facial UV maps. 

Due to the lack of 3D data, these methods have a 3D model based analysis-by-synthesis and compare input and rendered images. Since the rendered 3DMM images have only the facial region, a separate pretrained face segmentation network needs to be added to offset this gap. Thus, this will tie the network's performance to the effectiveness of the segmentation algorithm, and will also increase the computational load.  Furthermore, these methods are usually two-step training or cumbersome preprocessing. 

\begin{figure*}[!t]
\begin{center}
\centering\includegraphics[scale=0.43]{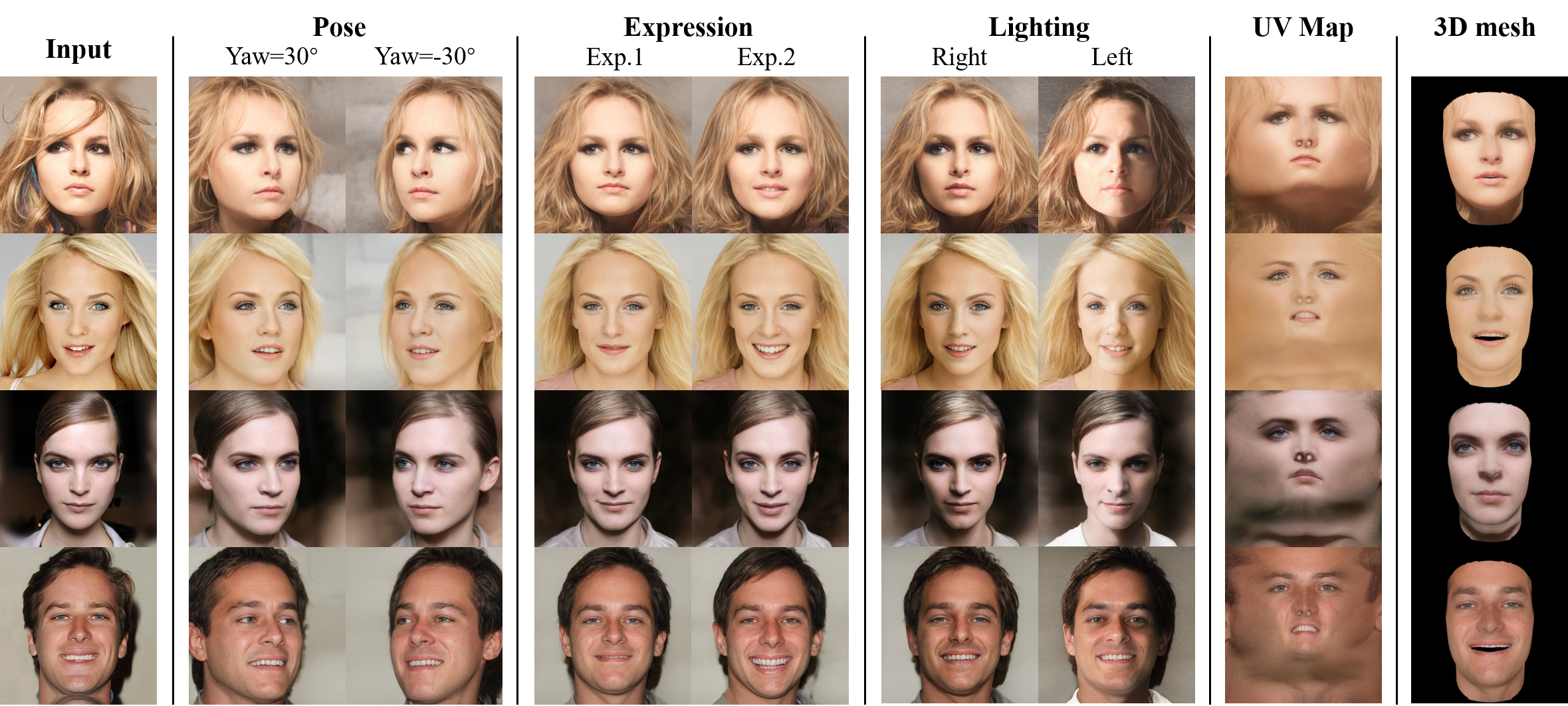}
\end{center}
\setlength{\abovecaptionskip}{0pt}
   \caption{The proposed method controls images generated by StyleGAN by applying semantic editing on corresponding 3D face meshes. As a by-product, we can also generate UV maps corresponding to the faces. The figure shows from left to right: original real image, face rotation, expression transformation, lighting, UV map and 3D mesh.}
\label{intro}
\end{figure*}

In this paper, we propose a novel approach capable of learning explicit control over facial pose, lighting, and expression in the latent space of StyleGAN via a self-supervised manner. Our model combines pre-trained StyleGAN and 3DMM, thus, we do not need any dataset and preprocessing. Our innovation lies on a network called 'Map and edit', which provides a direct transformation from the StyleGAN latent space to the 3DMM parameter space. After mapping the latent code to the 3DMM parameter space, we can edit the semantic parameters (pose, illumination and expression) of a face in 3DMM. Unlike in StyleGAN latent space, editing facial attributes in 3DMM allows more precise and explicit control. The inverse transform part of ‘Map and edit’ then allows the edited 3DMM to StyleGAN latent space, so that an edited image can be generated. In this way, our proposed model avoids the domain gap issue encountered by the previous methods. In addition, by combining synthesized multi-view face images and the corresponding visibility mask, our model also produces a high-quality UV map which would allow more detailed 3D mesh generation. Fig. \ref{intro} shows example results produced by our method.

Our contributions in this effort are:

\begin{itemize}
\item We designed a novel 3D prior guided 2D facial manipulation algorithm that achieves an effective and explicit facial attribute disentanglement and control.

\item The model can be trained in an end-to-end and self-supervised manner without any dataset and preprocessing.

\item We developed a novel scheme of generating high-fidelity face UV maps via our multi-view face image generator.

\item By transforming latent codes of the generated images into 3DMM for loss calculation between the generated image and 3DMM, our approach avoids the domain gap issue between the two spaces.

\end{itemize}

\section{Related Work}

\subsection{Attribute Editing in Latent Space}
Since its introduction, StyleGAN attracted much attention, particularly in semantic disentanglement of its latent space. In these efforts, supervised methods require off-the-shelf attribute classifiers or annotated images for specific attributes. InterfaceGAN \cite{shen2020interpreting} trains an SVM classifier to learn a boundary hyperplane for each binary attribute. Self-supervised learning methods, on the other hand, rely on externally pre-trained models or other models that can be used as labels to achieve training without using annotated images.  StyleClip \cite{Patashnik_2021_ICCV} and Clip2StyleGAN \cite{abdal2021clip2StyleGAN} creatively combines the generative power of a pre-trained StyleGAN generator with the vision-language model of CLIP, using text to guide the search for meaningful directions. Unsupervised methods, however, do not require pre-trained classifiers. Instead, GANspace \cite{2020ganspace} performs PCA on the latent space and discover editable and controllable semantic directions directly. Similarly, SeFa \cite{shen2021closedform} performs eigenvector decomposition on affine layers of StyleGAN. Some studies \cite{ZhuangICLR2021, voynov2020unsupervised} also found distinguishable semantic directions based on mutual information. Unfortunately, these latent space manipulation methods so far have difficulties in isolating and separating the target attributes from other properties. Additionally, most of these methods lack explicit attribute control.

\begin{figure*}[!t]
\begin{center}
\centering\includegraphics[scale=0.48]{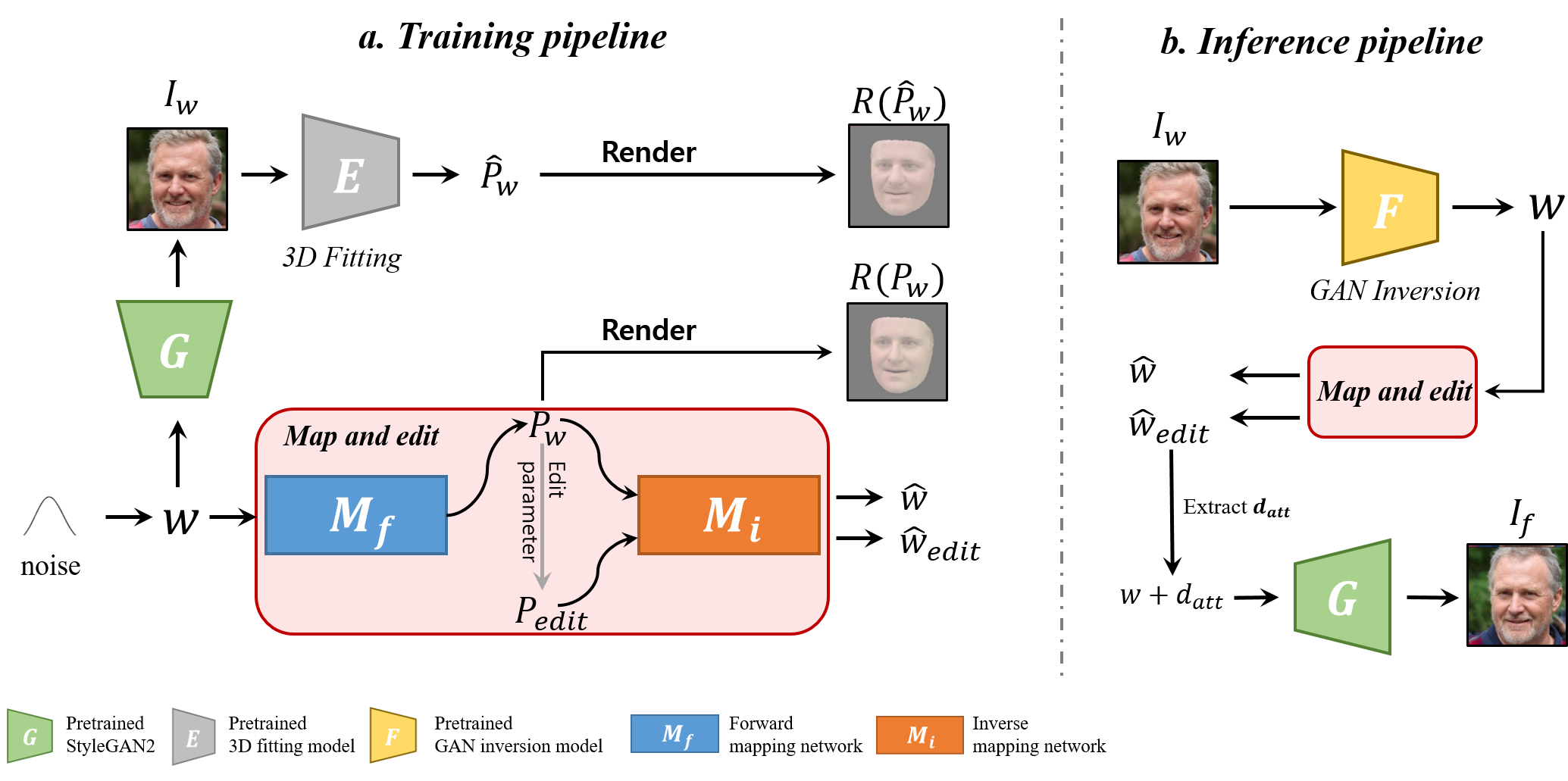}
\end{center}
\setlength{\abovecaptionskip}{0pt}
   \caption{Our model overview. \textbf{(a) Training pipeline}: we train the $M_f$ for mapping latent code $w$ to correspongding 3DMM parameter $P_w$. The $M_i$ is trained for mapping 3DMM parameter to latent code. \textbf{(b) Inference pipeline}: for real image editing, we first get the latent code $w$ via GAN inversion, then get paired latent code (${\hat{w}, \hat{w}_{edit}}$) via 'Map and edit'. Then extract $ d_{att}$, get the final image through $G(w+d_{att})$.}
\label{architecture}
\end{figure*}

\subsection{Face Rotation with 3D Prior Supervision}

Among the 3D face reconstruction methods, 3DMM was originally introduced by Blanz and Vetter \cite{blanz1999morphable}. They scanned 200 adult human head models in 3D, processed them by Principal Components Analysis (PCA), and built a parametric face model. Each coefficient vector can be regarded as different features of a human face, such as shape, expression, pose, texture, etc. The development of deep networks has given rise to more powerful nonlinear 3DMMs. Many methods \cite{tewari2017mofa, deng2019accurate, tran2019towards, tran2018nonlinear, koizumi2020look, ren2021pirenderer, DECA} use CNNs to project faces into the 3DMM space, and the process is followed by a differentiable renderer to synthesize reconstructed faces. Face rotation is one of the major tasks in facial editing, as it can be applied in face recognition and UV map completion. There are many recent studies \cite{9578771, zhou2020rotate, KowalskiECCV2020} that combine 3D prior and GAN to solve this problem. StyleRig \cite{tewari2020stylerig} performs the attribute manipulation by combining latent embeddings of StyleGAN and 3DMM parameters. DiscoFaceGAN \cite{deng2020disentangled} is also based on 3DMM, which promotes separation among face attributes through contrastive learning. MOST-GAN \cite{medin2022most} incorporates a 2D hair processing network, enabling extreme manipulation of lighting, facial expressions, and pose changes with fully separate 3D control over its physical properties. As these methods utilize 3DMM and use the generated faces in the loss function calculation, there is a domain gap between the rendered 3DMM faces and the generated faces. The domain gap often leads to a bias between the input image and the edited image. Additionally, most of these methods require two stages of training. Unlike an end-to-end training, it results greater complexity in the model learning process and may also may lead to a sub-optimal model.

Moreover, Neural Radiance Field (Nerf) \cite{Mildenhall20eccv_nerf, reiser2021kilonerf, Liu20neurips_sparse_nerf, Schwarz20neurips_graf,gu2021stylenerf} has received great attention in the field of 3D representation. Its key ideal is to non-explicitly model a complex static scene with a neural network. $\pi$-GAN \cite{Chan20arxiv_piGAN} achieves SOTA in the study of unsupervised multi-view face generation using GAN and Nerf. However, this method is more difficult to edit in real images. \cite{kwak2022injecting} used a Nerf-based 3D face generator for synthesizing images and pseudo labels as facial poses are varied. But it cannot edit the geometry and lighting of the face.

\subsection{Facial UV Map Completion}

Traditional 3DMMs use parametric models to generate face UV maps. UV map helps to distribute 3D texture data onto 2D planes and provides common per-pixel alignment for all textures. The limitation of these methods is that the generated textures are of lower quality and less realistic. UV map completion using deep neural networks has been actively researched in the past few years \cite{deng2018uv, yin2021weakly, lin2021meingame, na2020facial, lee2020uncertainty, lattas2020avatarme, 9442802}. GANFIT \cite{gecer2019ganfit} utilizes GANs to train a facial texture generator in a UV space. DSD-GAN \cite{kim2021learning} applies two discriminators in a UV map space and an image space to simultaneously learn face structure and texture details. Although these methods can generate realistic UV maps, they require extensive preprocessing or building a UV map dataset. OSTeC \cite{gecer2021ostec} combines StyleGAN and 3DMM and uses an optimization method to generate UV maps with only one image. However, each image requires a heavy computational load, and often fails to maintain personal identity of input image.

\section{Methods}
Our approach combines 3DMM face models with pre-trained StyleGAN to achieve attribute-controlled and disentangled 2D face models. We propose a ‘Map and edit’ that can map directly each other in $W$ space and 3DMM coefficient space, and edit the images generated by 2D StyleGAN by editing the properties of the 3DMM coefficient corresponding to the latent code. The proposed algorithm allows us to extrapolate beyond what is well represented in the training set, allowing precise control over the pose, facial expression, and lighting of the face. In addition, since we can generate face images at specified angles, we can take advantage of this for multi-view face image generation. Then, perform realistic 3D facial texture completion and generate the corresponding 3D mesh.

\subsection{Map and Edit Network}
Our face image manipoluation method relies on pretraind StyleGAN and pretrained 3DMM fitting model. StyleGAN $G$ can be seen as a function that maps latent codes $w \in \mathbb{R}^{18\times512}$ realistic portrait images of human faces $I_w = G(w) \in \mathbb{R}^{3\times W \times H}$. For a given 2D face image $I$, our method relies on estimating the 3D reconstruction of the face by using a CNN model $E$. Therefore, we fit an off-the-shelf  algorithm to estimate its geometry and color based on a set of 3DMM parameter $x=(\alpha, \beta, \delta, \gamma, \phi, t) \in  \mathbb{R}^{257}$, which depict the identity $\alpha \in  \mathbb{R}^{80}$, texture$ \beta \in  \mathbb{R}^{80}$,  expression $\delta \in  \mathbb{R}^{64}$, illumination $\gamma \in  \mathbb{R}^{27}$, face rotation $\phi \in SO(3)$, and translation $t \in  \mathbb{R}^3$.

Our core idea is to obtain the corresponding semantic direction of the StyleGAN's latent space by modifying the 3DMM semantic parameters and apply it to the original latent code. We propose a ‘Map and edit’ network consisting of two MLP networks $M_f$ and $M_i$ that can map each other between the latent space and the 3DMM parameter space, $i.e.$ $M_f:\mathbb{R}^{18\times512}\Rightarrow \mathbb{R}^{257}, M_i:\mathbb{R}^{257}\Rightarrow \mathbb{R}^{18\times512}$. It allows us to perform similar editing of the facial images generated by StyleGAN based on semantic and interpretable control parameters. As shown in Figure \ref{architecture} (a), we first randomly sample $w$ and generate the image $I_w=G(w)$ through the StyleGAN generator. Then, the 3DMM parameters $\hat{P}_w=E(I_w)$ of $I_w$ are extracted using the pretrained 3D fitted model. The forward mapping network $M_f$ converts the latent code $w$ into $P_w=M_f(w)$, and then generates the reconstructed latent code $\hat{w}=M_i(P_w)$ through the inverse mapping network $M_i$. At this time, we edit the semantic control parameters of $P_w$ (expression $\delta$, face rotation $\phi$ and illumination $\gamma$) to get $P_{edit}$ after the semantic information has been changed, and then use $M_i $ maps them to the $W$ space to get $\hat{w}_{edit}=M_i(P_{edit})$, where $\hat{w}_{edit}$ is the semantic-transformed version of $\hat{w}$.

To train ‘Map and edit’, we generate paired 3D meshes with 3DMM parameters $P_w$ and $\hat{P}_w$, and use a differentiable renderer $R$ to generate the corresponding rendered images $R(P_w)$ and $R (\hat{P}_w)$. We train $M_f$ by minimizing the discrepancy of ($R(P_w), R(\hat{P}_w)$) and ($P_w, \hat{P}_w$) respectively, by minimizing the discrepancy of $w$ and $\hat{ w}$ to train $M_i$. Details will be explained in the loss functions section.

\subsection{Semantic Direction Extraction}

\begin{figure}[!t]
\begin{center}
\centering\includegraphics[scale=0.45]{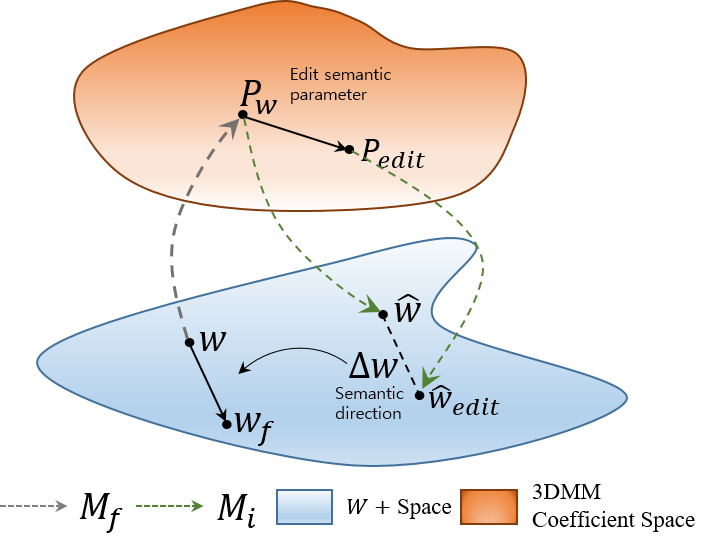}
\end{center}
\setlength{\abovecaptionskip}{0pt}
   \caption{Illustration of attribute editing in our method.}
\label{inference}
\end{figure}

As mentioned above, we obtain $\hat{w}$ and $\hat{w}_{edit}$ using the ‘Map and edit’ network. In ideal case, $\hat{w}_{edit}$ is our edited version of latent code. However, one of the key issues is that the dimension of the latent code ($18\times512$) is much larger than the dimension of the 3DMM parameter ($257$). This will lead to the loss of the some information of $w$, that is, the $w$ and $\hat{w}$ will produce a certain degree of gap in identity characteristics. Thus, it cannot use the $w_{edit}$ to generate the final version image directly.

To address the problem of information loss caused by dimensional differences, we propose a simple but effective method. As shown in Fig. \ref{architecture} (b), in the inference stage, a pretrained GAN inversion model \cite{tov2021designing} is employed to embedding an input image $I_w$ into $w$. Through ‘Map and edit’, we obtain paired latent codes ($\hat{w}, \hat{w}_{edit}$), and extract the corresponding semantic direction $d_{att}$. Then, based on the semantic direction $d_{att}$ which is extracted by explicitly controlling the parameters of the 3DMM. To describe process of semantic direction extraction in more detail, it is shown in Fig. \ref{inference}. We first obtained $P_w=M_f(w)$, and modify the corresponding semantic parameter to obtain $p_{edit}$. At this time, $\hat{w}$ and $\hat{w}_{edit}$ with the same identity attribute but different semantic direction are obtained through $M_i$. $\hat{w}$ and $\hat{w}_{edit}$ have the same identity characteristics except for the attributes we need to change. Benefit from the disentanglement of the StyleGAN's latent space, we can effortlessly extract the semantic direction $d_{att}$ with the following equation: $d_{att} = \hat{w} - \hat{w}_{edit}$. Finally, we utilize this semantic direction $d_{att}$ to edit the $w$ and produce the final image $I_f=G(w+d_{att})$.

\subsection{UV Map Completion}

With the powerful image generation capabilities of GANs, we can use StyleGAN as a prior model to complete unseen regions generated from a single image.  Recently, \cite{gecer2021ostec} proposed a method which is requires iterative optimization of several images with target angles and progressively refines the UV map. Although this method achieves good results on high quality UV map generation. This leads to the problem of inconsistent identities of the generated face images and the time-consuming problem of optimizing each input image. In contrast, our method can explicitly generate multi-view face images from target angles using pretrained StyleGAN. Finally, we combine them by blending and obtain a high quality UV map. Notably, our method has superior generalization, and the process are simpler and faster when synthesizing UV maps. For more details, please refer to the appendix.


\subsection{Loss Functions}

We design a loss function to train the ‘Map and edit’ network. Basically we can divide the training into two goals, one is from a network $M_f$ that can map the latent space to the 3DMM parameter space, and the other is a network $M_i$ that maps backwards. Within the training loop, we design five types of loss functions, i.e. the rendered image loss, the 3DMM parameter loss,  the latent  reconstruction loss, landmark loss and the regularization loss. The final loss is simply represented as:
\begin{equation}
    L_{final} = L_{ren}+\lambda_{p}L_{p}+\lambda_{lat}L_{lat}+\lambda_{lm}L_{lm}+L_{reg}
\end{equation}

\subsubsection{Rendered Image Loss}
We first design a rendered image loss for training $M_f$ that can directly map from latent space to 3DMM space. In consequence, we calculate the distance of the rendered 3DMM image as follow:
\begin{equation}
    L_{ren} = ||R(P_w)-R(\hat{P}_w) ||_1
\end{equation}
where $P_w$ and $\hat{P}_w$ are 3DMM parameters obtained by $M_f(w)$ and $E(I_w)$, respectively. And, $R(\cdot)$ denotes the differentiable renderer.

\subsubsection{3DMM Parameter Loss}
Although it is possible to learn rough information about pose and skin color and lighting through $L_{map}$. However, the details of texture and expression are not well-reacted. Therefore, we add the loss of 3DMM parameter reconstruction, called $l_p$, to increase the response to these details in 3DMM space.
\begin{equation}
    L_{p} = ||P_w-\hat{P}_w ||_1
\end{equation}

\subsubsection{Latent Reconstruction Loss}
To ensure that it is possible to go from 3DMM space to latent space, the latent reconstruction loss is employed as follow:  
\begin{equation}
    L_{lat} = ||w-\hat{w} ||_1
\end{equation}
where $\hat{w}=M_i(M_f(w))$ is denote the reconstruction version of $w$.

\subsubsection{Landmark Loss}
We also use landmarks as supervision for expression and pose. We do not extract landmarks directly at the image level, but utilize landmark index from vertices of 3DMM.
\begin{equation}
    L_{lm} = 1/N\sum^{N}||q_w - \hat{q}_w ||_1 +                     1/N\sum^{N}||q_{edit} - \hat{q}_{edit} ||_1
\end{equation}
where $q_w$, $\hat{q}_w$ and $q_{edit}$ are landmarks extract from $P_w$, $\hat{P}_w$ and $P_{edit}$. And $N$ is the number of landmarks, we use 68 points in our study. Since we do not have ground truth for $\hat{w}_{edit}$. Thus, we add the second term in landmark loss to ensure that the attribute of $\hat{w}_{edit}$ is consistent with the attribute of $P_{edit}$. $\hat{P}_{edit}$ is the reconstruct version of $P_{edit}$.

\subsubsection{Regularization Loss}
To prevent face shape and texture degeneration, we utilize a commonly used loss to accelerate the convergence of the 3DMM parameter similar to \cite{deng2019accurate}:
\begin{equation}
    L_{reg} =  \lambda_\alpha||\alpha||^2 + \lambda_\beta||\beta||^2 + \lambda_\delta||\delta||^2
\end{equation}
where $\alpha$, $\beta$ and $\delta$ are identity, expression and texture parameters of 3DMM.

\section{Experimental Results}

This section presents qualitative and quantitative comparisons with state-of-the-art methods \cite{deng2020disentangled, KowalskiECCV2020, shi2021lifting, kwak2022injecting} and an analysis of our method. Additional experimental results, model details, and detailed training methods can be found in Supplementary Materials.

\section{Implementation Details}

In this paper, we explore disentangled manipulations in the latent space $W$ of pretrained StyleGAN2 model \cite{karras2020analyzing}. The StyleGAN2 is trained on FFHQ dataset \cite{karras2019style} with a Pytorch version. We use mesh rasterizer of Pytorch3D \cite{ravi2020pytorch3d} for differentiable rendering. We utilize the pretrained model of \cite{tov2021designing} and \cite{deng2019accurate} as GAN inversion and 3DMM fitting networks. We use MTCNN face detector \cite{zhang2016joint} extract 5-point facial landmark for 3DMM fitting. We sample $100k$ latent codes $w\in \mathbb{R}^{18\times512}$ and generate the corresponding images $I_w=G(w)$ at the resolution of $1024\times 1024$. The resolution of UV map is also $1024\times1024$. Our model is trained entirely ent-to-end and simultaneously can train three controllable attributes (pose, lighting and expression). During training process, we optimize parameters using Adam with $0.1$ learning rate. Due to memory constraints, we set the batch size to 2. And we set our balancing factors as the following: $\lambda_{p}=1.0$, $\lambda_{lat}=1.0$, $\lambda_{lm}=0.01$. The fitting converges in around 100,00 iterations on an Nvidia RTX 2080 Ti GPU. More training details can be found in the supplementary material.

\subsection{Manipulation Results}

\begin{figure*}[!t]
\begin{center}
\centering\includegraphics[width=\textwidth]{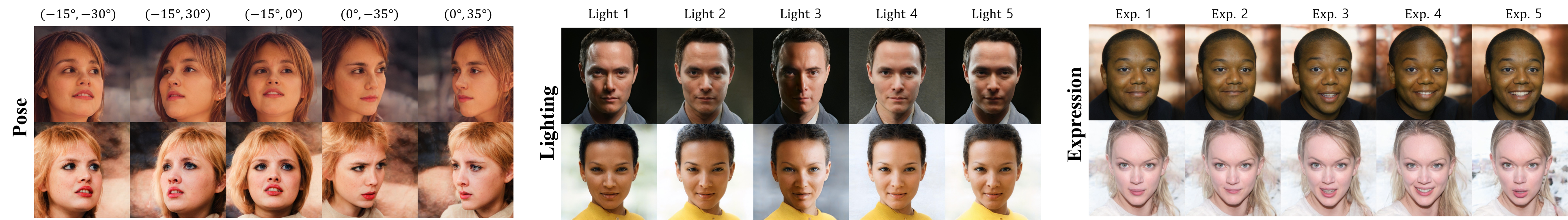}
\end{center}
\setlength{\abovecaptionskip}{0pt}
   \caption{Face images manipulation results of our method. As shown in the figure, our method can control the pose, lighting and expression of the generated image by means of explicitly control.}
\label{exp1}
\end{figure*}

\begin{figure}[!t]
\begin{center}
\centering\includegraphics[scale=0.3]{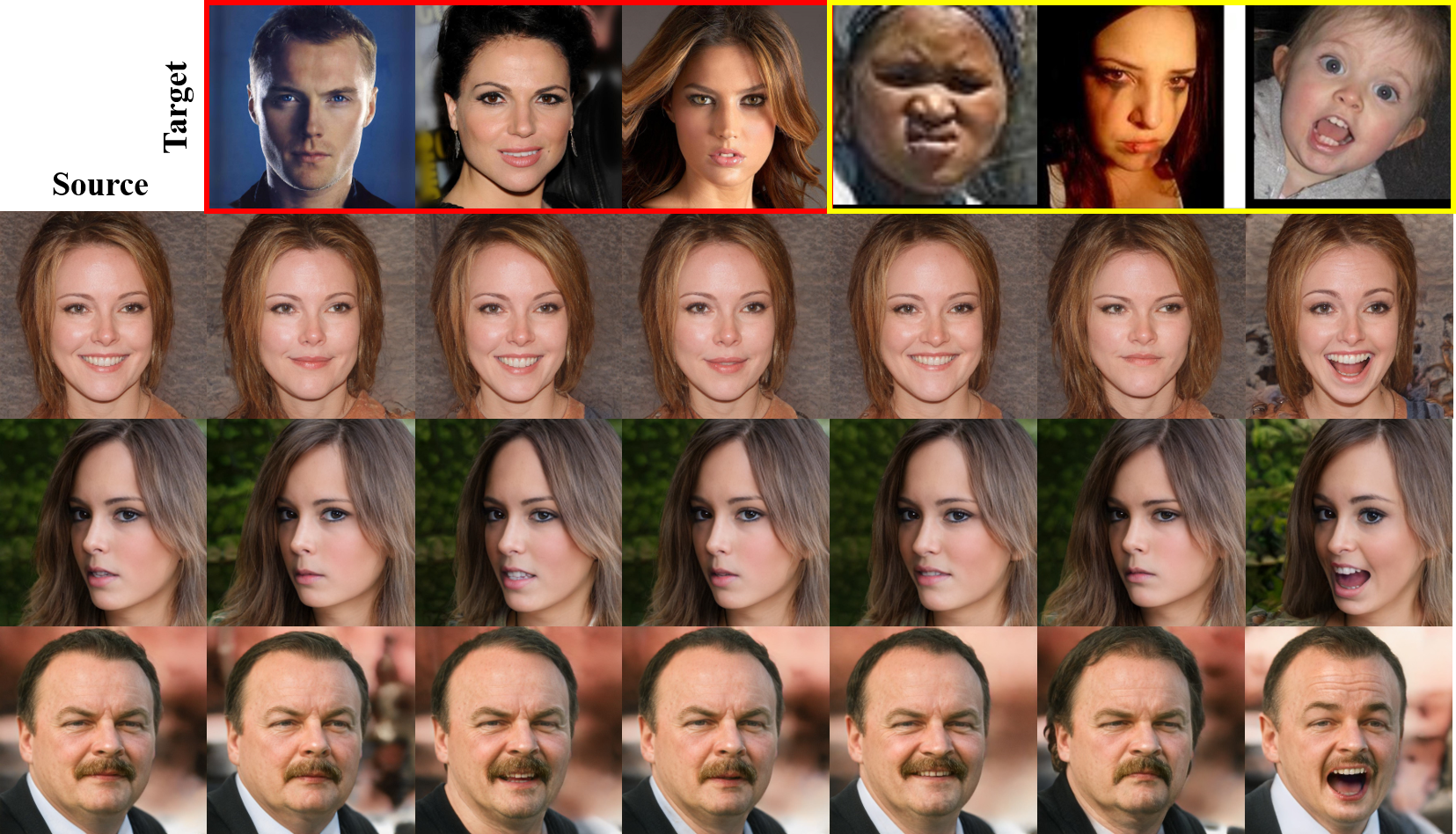}
\end{center}
\setlength{\abovecaptionskip}{0pt}
   \caption{Based on the reference generation results, we utilize $M_f$ to extract the expression semantic of real images and combine them with the source images from Celeba-HQ. The reference image in the red box is from Celeba-HQ, and the reference image in the yellow box is from the real world.}
\label{exp3}
\end{figure}

As shown in Fig. \ref{exp1}, We first image manipulation results of our method. Explicit control over the 3D parameters allows us to turn StyleGAN into a conditional generative model. We can simply enter pose, expression or lighting parameters into $'$Map  and edit$'$ to generate an image corresponding to the specified parameters. That is, we have explicit control over the pretrained generative model. Furthermore, changes in pose, expression and lighting are highly disentangled. Fig. \ref{exp3} shows that we can generate images of corresponding expressions from the expression properties of reference images. We edit the image by extracting the expression parameters of 3DMM from the reference image and extracting the corresponding expression direction.

\begin{figure}[!t]
\begin{center}
 \centering\includegraphics[scale=0.27]{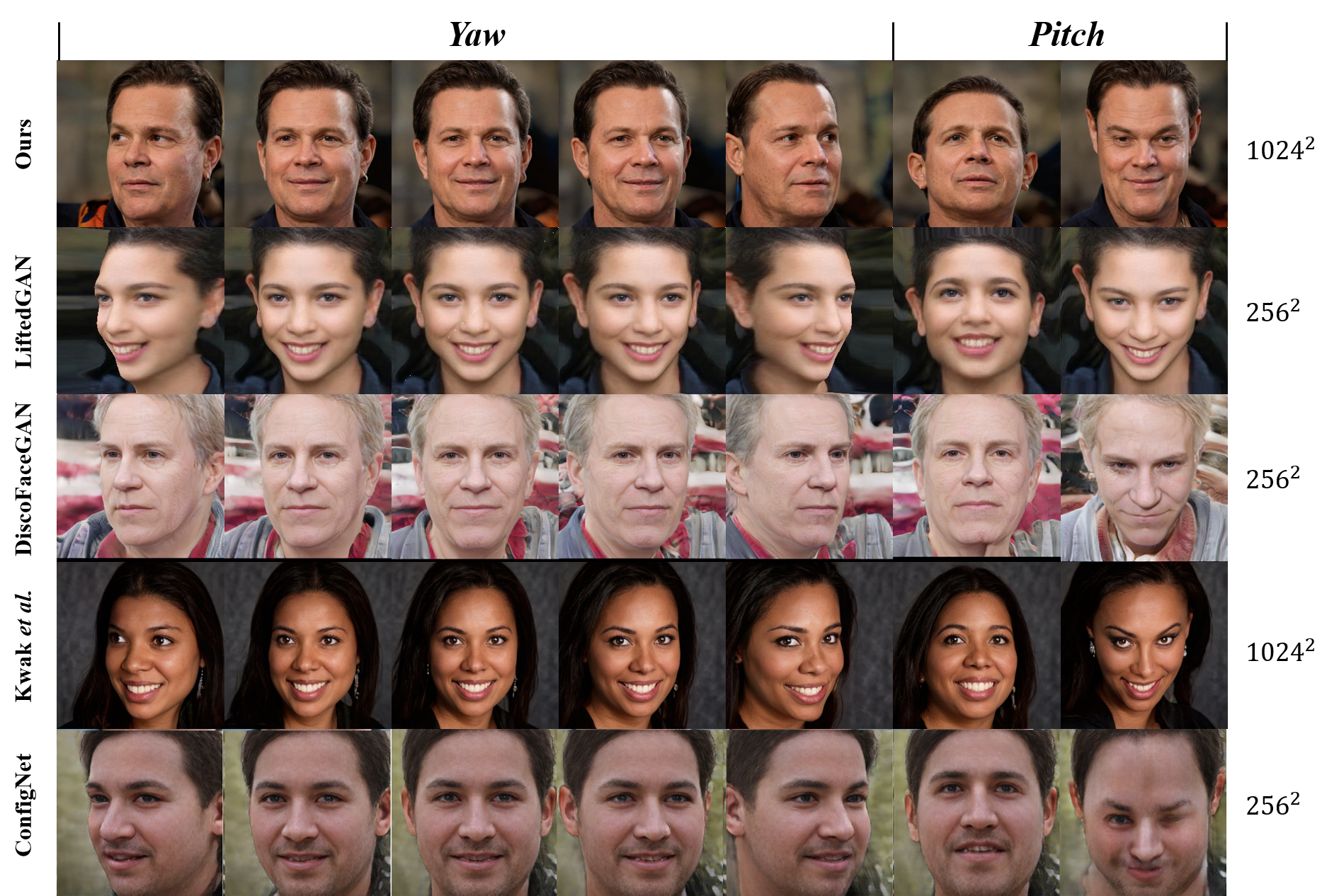}
\end{center}
\setlength{\abovecaptionskip}{0pt}
   \caption{Qualitative comparison results of random face rotation. It is clear that our model can rotate high-resolution images and maintain the identity of the original image. Among them, DiscoFaceGAN cannot do explicit control.}
\label{exp2}
\end{figure}

\begin{figure}[!t]
\begin{center}
\centering\includegraphics[scale=0.38]{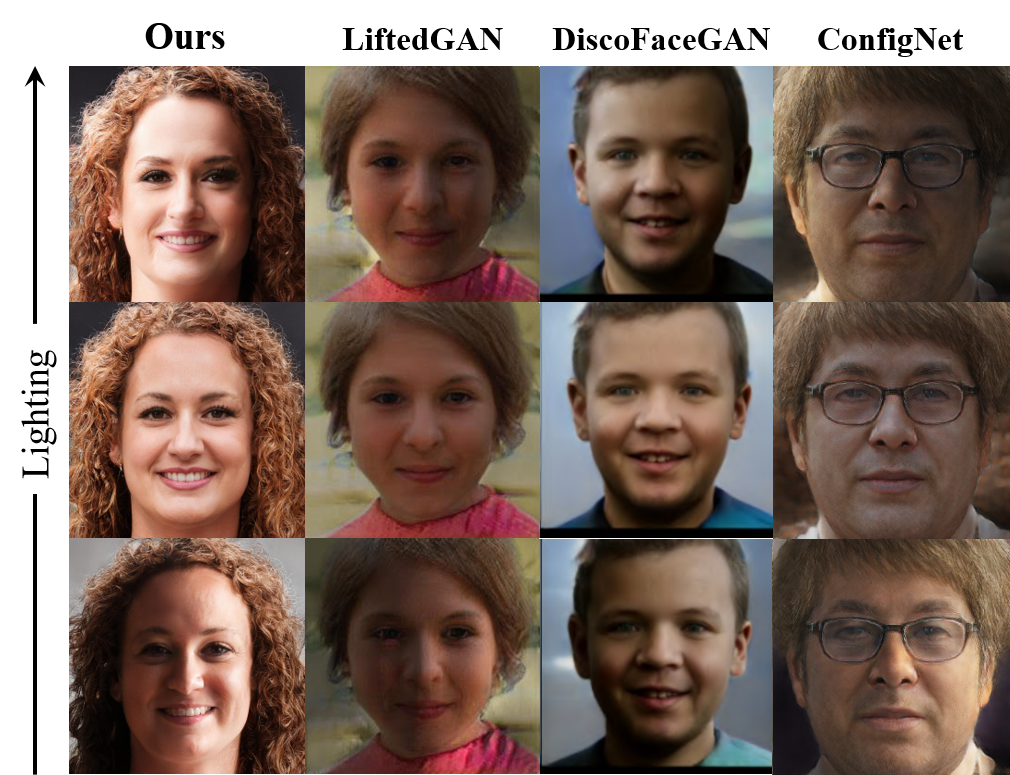}
\end{center}
\setlength{\abovecaptionskip}{0pt}
   \caption{Qualitative comparison results of relighting manipulation.}
\label{exp4}
\end{figure}

Fig. \ref{exp2} shows the comparison results of our method, discofacegan, configNet, Kwak et al and LiftedGAN on pose variation. The results show that our method is better at rotating portrait images while keeping other conditions such as identity unchanged. Result of LiftedGAN is distorted after a large angle rotation. The result of ConfigNet is distorted in the rotation result in the pictch direction. DiscofaceGAN and Kwak \etal{} have achieved good results in this regard. Whereas DiscofaceGAN relies on training the entire GAN, Kwak \etal{} relies on nerf-based generators. These all require costly training resources. And DiscofaceGAN cannot explicit control the face pose. Our method only needs to optimize the proposed ‘Map and edit’ edit’ network and does not require any dataset. Figure \ref{exp4} shows a comparison of relighting. It can be seen that we can freely change the lighting direction and intensity. It is clear and our method can relight the resulting image with much greater light intensity than other baselines.

\subsection{Quantitative Evaluation}

For quantitative evaluation, we use FID \cite{heusel2017gans} score, pose accuracy estimated by 3D model \cite{guo2020towards}, and identity similarity \cite{deng2019arcface} as evaluation metrics. In order to evaluate the quality of the image generation, we use the randomly generated images of 50K and real image of 50K to calculate the FID. For pose error and identity similarity calculation, we generated 2.5k images on each angle. 

As shown in Tab. \ref{pose_error} and Tab. \ref{consine_similarity}, compared to other baseline, our method achieves competitive scores in pose accuracy and provides superior results in terms of visual quality and identity consistency. In the FID comparison results, since the pretrained StyleGAN\cite{karras2020analyzing} is also used, the results of \cite{kwak2022injecting} are similar to ours. 

\begin{table}[htbp]
 \setlength{\tabcolsep}{0.5mm}
  \centering
  \caption{Quantitative comparison on FID ($\downarrow$) and pose error ($\downarrow$) with random generated face.}
  \begin{tabular}{*{5}{c}}
      \hline
      Method & ConfigNet & LiftedGAN & Kwak \etal{} & Ours  \\
      \hline
      FID & 33.41 & 29.81 & \textbf{4.72} & 4.79 \\
      Pose error ($\times10^{-2}$) & 9.56 & 5.52 & 4.24 & \textbf{3.26} \\
      \hline
  \end{tabular}
\label{pose_error}
\end{table}

\begin{table}[htbp]
 \setlength{\tabcolsep}{0.8mm}

  \caption{Cosine similarity ($\downarrow$) comparison results of face rotation at specific angles with random generated face.}
  \begin{tabular}{*{9}{c}}
      \hline
      Yaw & $-40^{\circ}$ & $-30^{\circ}$ & $-20^{\circ}$ & $-10^{\circ}$ & $10^{\circ}$ & $20^{\circ}$ & $30^{\circ}$ & $40^{\circ}$ \\
      \hline
      Kwak \etal{} & 0.60 & 0.73 & 0.83 & 0.94 & 0.94 & 0.83 & 0.72 & 0.61 \\
      ConfigNet & 0.15 & 0.29 & 0.43 & 0.71  & 0.72 & 0.47 & 0.29 & 0.15 \\
      LiftedGAN & 0.28 & 0.43 & 0.57 & 0.82  & 0.82 & 0.58 & 0.42 & 0.29 \\
      Ours & \textbf{0.68} & \textbf{0.79} & \textbf{0.89} & \textbf{0.96} & \textbf{0.96} & \textbf{0.90} & \textbf{0.80} & \textbf{0.69}\\
      \hline
  \end{tabular}
  
   \setlength{\tabcolsep}{2.04mm}
    \begin{tabular}{*{7}{c}}

      Pitch  & $-30^{\circ}$ & $-20^{\circ}$ & $-10^{\circ}$ & $10^{\circ}$ & $20^{\circ}$ & $30^{\circ}$  \\
      \hline
      Kwak \etal{} & \textbf{0.72} & 0.84 & \textbf{0.95} & 0.94 & 0.85 & \textbf{0.75} \\
      ConfigNet & 0.28 & 0.41 & 0.72 & 73 & 0.42 & 0.28  \\
      LiftedGAN & 0.42 & 0.61 & 0.84 & 0.84 & 0.63 & 0.42 \\
      Ours & \textbf{0.72} & \textbf{0.85} & \textbf{0.95} & \textbf{0.97} & \textbf{0.88} & \textbf{0.75} \\
      \hline
  \end{tabular}
\label{consine_similarity}
\end{table}

\subsection{Ablation Study}
\begin{figure}[!t]
\includegraphics[scale=0.37]{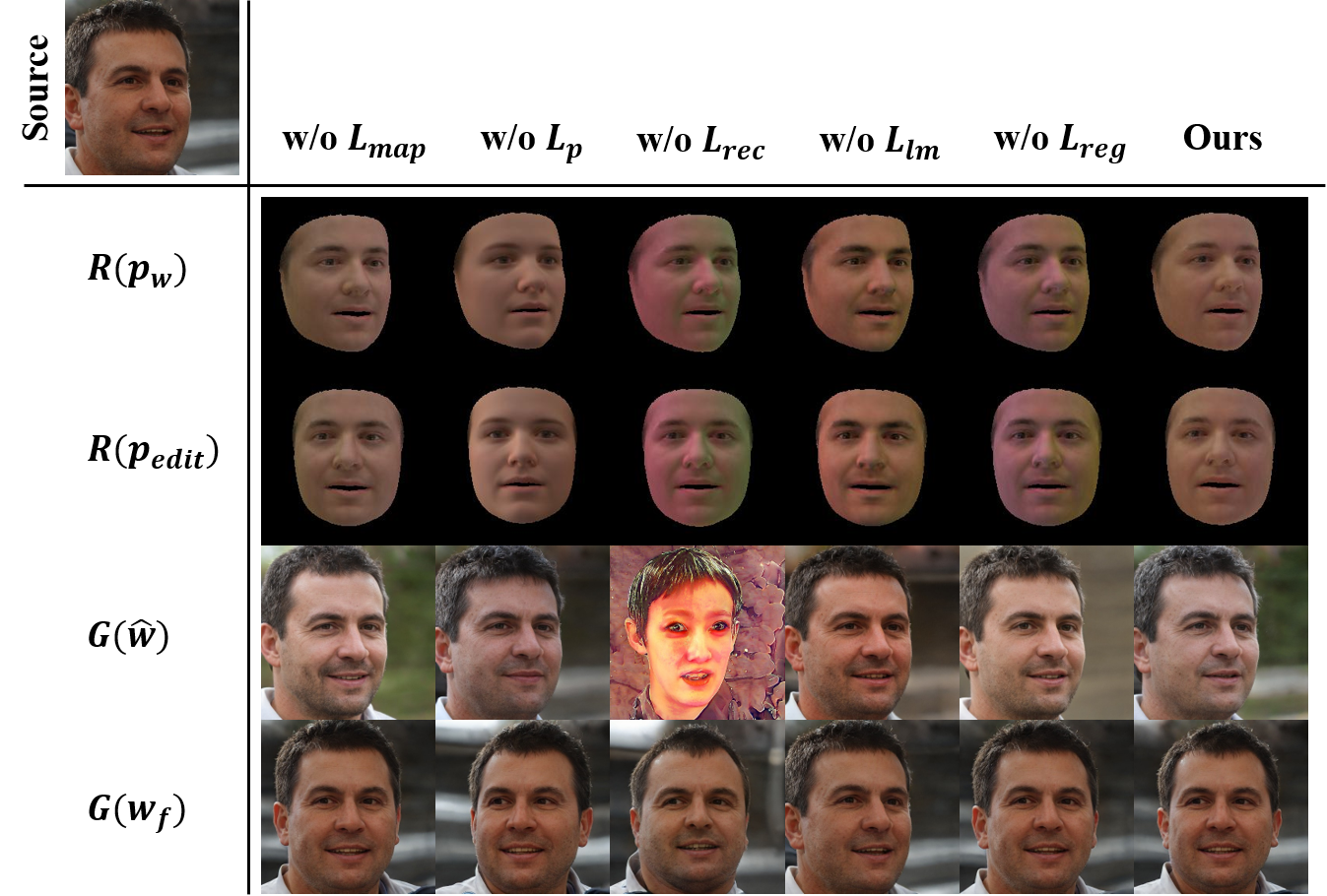}
\caption{Qualitative ablation study results.}
\label{ablation}
\end{figure}

To verify the effectiveness of our proposed mixed-level loss function, we train our model with different losses. Some typical results are shown in Fig. \ref{ablation}. The combination of rendered image loss and p loss can better help to map the latent code reasonably to the 3DMM space. When we remove the rendered image loss, the color and lighting of $R(P_w)$ are biased. When we remove the p loss, the texture details of the $R(P_w)$ to be missing. The latent reconstruction loss helps map the 3DMM coefficients $p$ back to the $W$ latent space. The landmark loss can help transform the geomtry changes (pose and expression) from $P_w$ to $P_{edit}$. The regularization loss helps the $M_f$ network to speed up the convergence. It can be seen from the results of the last line that without the rendered image loss and regularization loss, the face rotation does not have much impact. However, due to the poor feedback on the light in these two cases, a large error will occur when adjusting the lighting.

\subsection{UV Map Completion Results}
\begin{figure}[!t]
\begin{center}
\centering\includegraphics[scale=0.19]{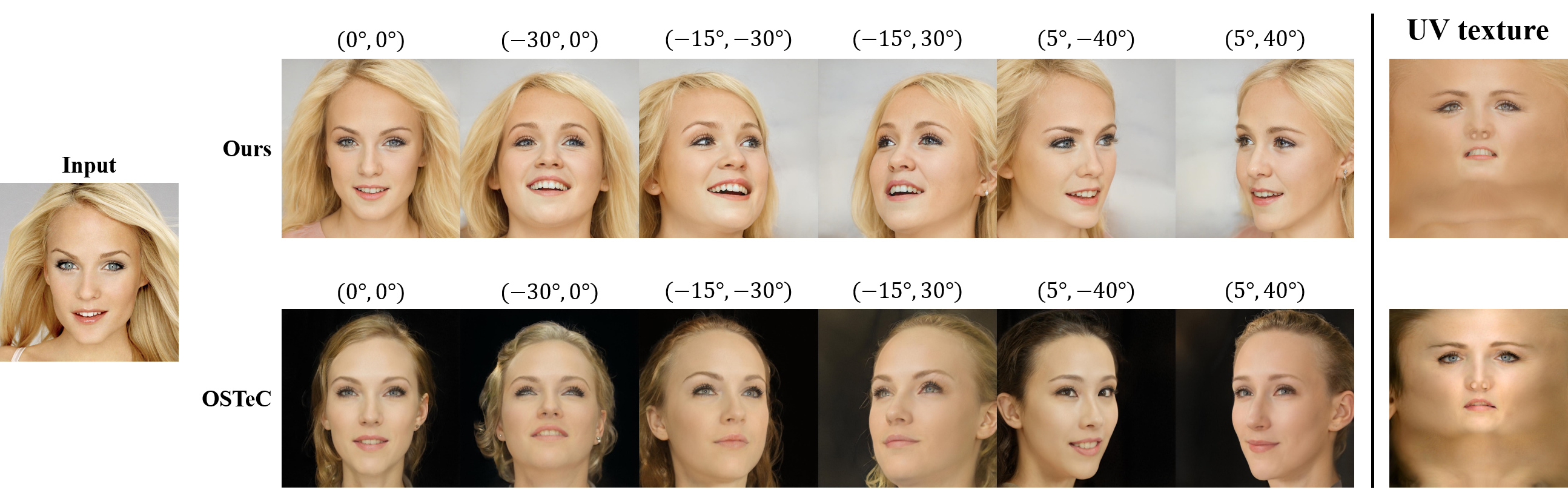}
\end{center}
\setlength{\abovecaptionskip}{0pt}
  \caption{Comparison results with rotations of real face images from OSTeC. The right column is the UV map synthesized by the corresponding method.}
\label{uv_exp1}
\end{figure}

\begin{figure}[!t]
\begin{center}
\centering\includegraphics[scale=0.32]{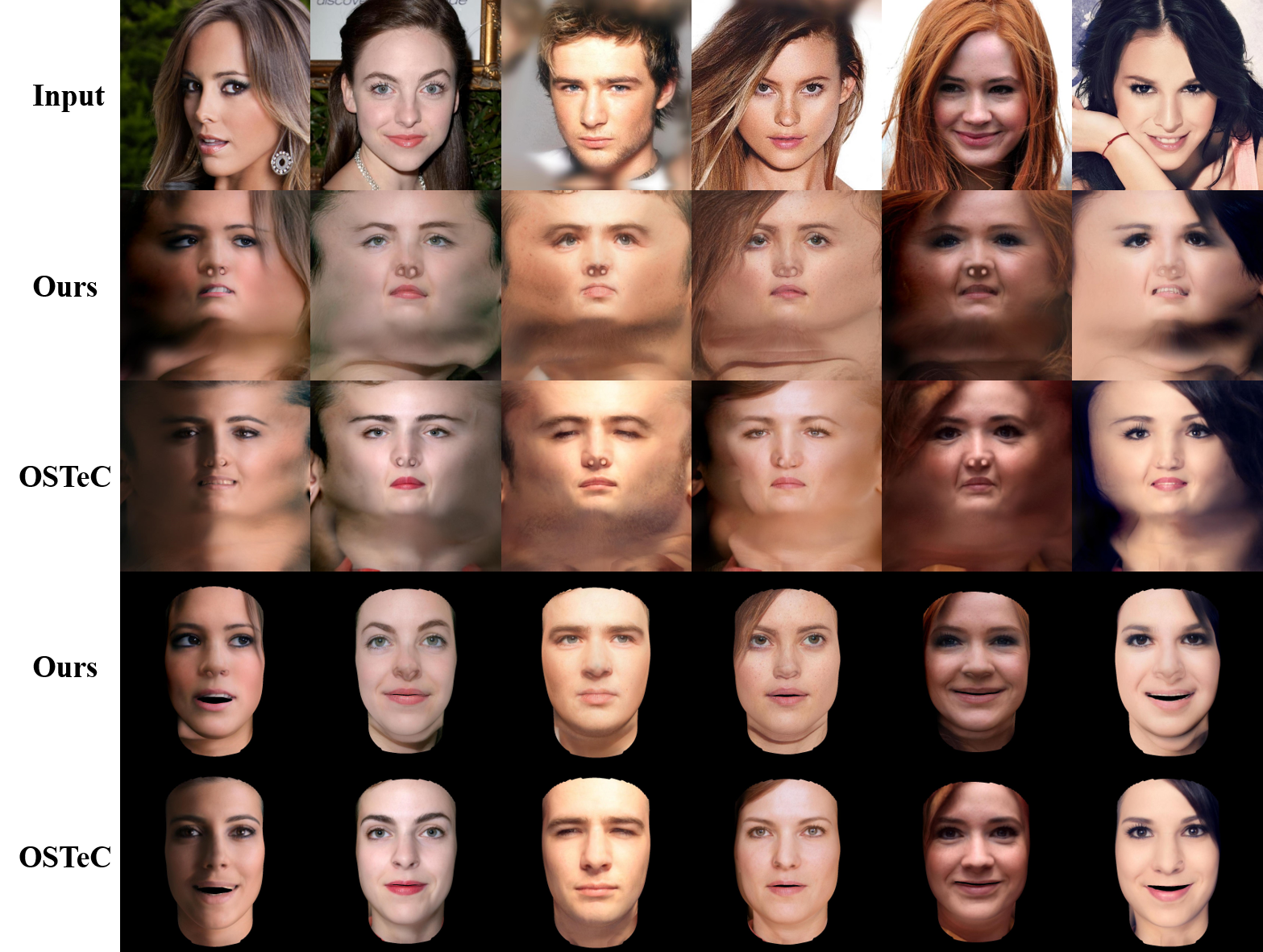}
\end{center}
\setlength{\abovecaptionskip}{0pt}
  \caption{UV map completion of real images with OSTeC and corresponding 3D mesh synthesis results. The second/third rows, and fourth/fifth rows are the UV map and rendered 3D mesh results, respectively.}
\label{uv_exp2}
\end{figure}

Many 3D texture completion methods often rely on large-scale high-quality 3D appearance data, which is expensive and difficult to collect. With our model, a high-quality face UV textue is synthesized by generating a specified multi-view face combined with a visibility mask as a by-product of our method. More recently, OSTeC \cite{gecer2021ostec} uses a model-optimized approach that can achieve face UV map synthesis without using any 3D data. However, they need to be optimized for each input image, which requires a long inference time. It can be seen from the fig. \ref{uv_exp1} that the identity of the multi-view face image generated by OSTeC is quite different from that of the input image, and the identity consistency of each angle of the face is relatively low. In contrast, the face images generated by our method maintain better identities. Fig. \ref{uv_exp2} shows more UV map completion comparison results for CelebA-HQ \cite{karras2017progressive}. Obviously, our results better preserve the details of the source image, such as skin color, facial expression and gaze position, etc.

In addition to this, we compared the runtime from single image to UV map generation with OSTeC in our environment. Due to time constraints, we tested OSTeC and our model with 50 and 150 images respectively. Table \ref{uv_table} shows our average run time is about $30 \%$ of OSTeC. The main time-consuming difference is in the multi-view face generation part, since our model is based on a generator with better generalization, i.e., pretraiend StyleGAN. Therefore, this part of the time consumption can be almost ignored in our method.


\begin{table}[htbp]
 \setlength{\tabcolsep}{6mm}
  \centering
  \caption{Comparison of runtime from single image to UV map generation.}
  \begin{tabular}{*{3}{c}}
      \hline
      Method & Ours & OSTeC \\
      \hline
      Run time $\downarrow$ (Sec.) & \textbf{162.32} & 533.89 \\
      \hline
  \end{tabular}
\label{uv_table}
\end{table}


\section{Conclusion}

We propose a novel method to provide face-like pose, expression, and lighting control on a pretrained StyleGAN network. Since our network is trained in a self-supervised and end-to-end manner, no training data is required. Our method allows explicit control of generated image semantics via corresponding 3DMM parameters. The results demonstrate that our method disentangles facial pose, expression, or lighting more effectively compared to the other methods. By generating visually salient, accurate and identity consistent face UV maps, our method delivers highly realistic images.

\clearpage
\bibliography{aaai22}
\end{document}


\maketitle

\section{Implemetation Details}
\subsection{Training Details}
Our goal is to reflect the semantic parameter transformation (pose, lighting and expression) of the 3DMM \cite{bfm09} to the corresponding latent code. To fortify disentanglement, we enforce the invariance of the latent representations for image generation in a contrastive manner: we vary one latent variable while keeping others unchanged, and enforce that the difference on the generated face images relates only to that latent variable. In the description of this paper, we obtain the disentanglement ability through $P_{edit}$. In the training phase, we randomly select one of the three attributes of $P_w$ for editing each iteration. We follow the following principles to get $P_{edit}$:  $\textbf{(1)}$ \textbf{Pose:} When editing the parameters of pose, we randomly generate angles for yaw and pitch from a uniform distribution to replace the original angle information. $\textbf{(2)}$ \textbf{Lighting:} Randomly sample from a uniform distribution and replace the original lighting parameter. $\textbf{(3)}$ \textbf{Expression:} Expression-related parameters consist of 64 numbers, which are extracted and sorted by PCA. In other words, the parameters in the front have a greater impact on the expression transformation, thus, we only choose to edit the first 30 parameters. When editing parameters, we randomly pick two and randomly sample from a uniform distribution in place of the original parameters. 

\subsection{Inference Details}
According to the properties of StyleGAN's 'style mixing', that is, different channels of $w$ are responsible for different $styles$. Therefore, after extract the semantic direction $d_{att}\in \mathbb{R}^{18\times512}$, we only use the part of channls of $d_{att}$ for attribute editing: \textbf{(1) pose:} 0-3; \textbf{(2) expression:} 4-6;  \textbf{(3) lighting:} 8-9. 

\begin{figure}[!t]
\begin{center}
\centering\includegraphics[scale=0.45]{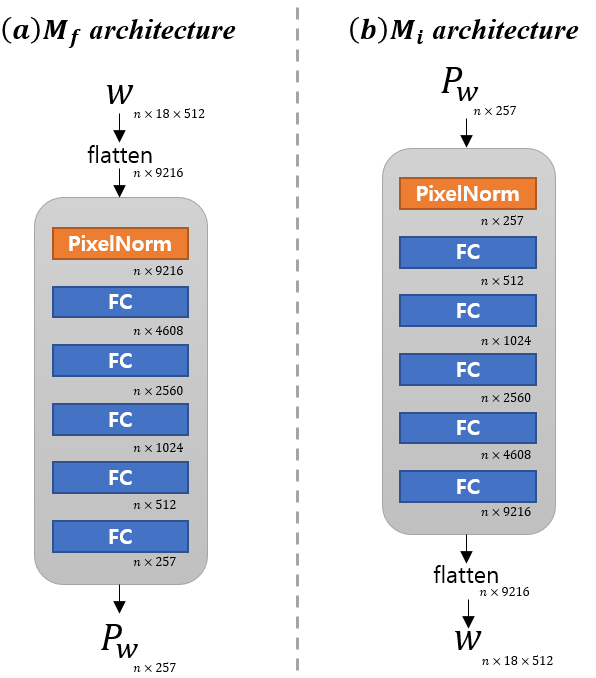}
\end{center}
\setlength{\abovecaptionskip}{0pt}
   \caption{Architecture of 'Map and edit' network.}
\label{mlp}
\end{figure}

\begin{figure}[!t]
\begin{center}
\centering\includegraphics[scale=0.27]{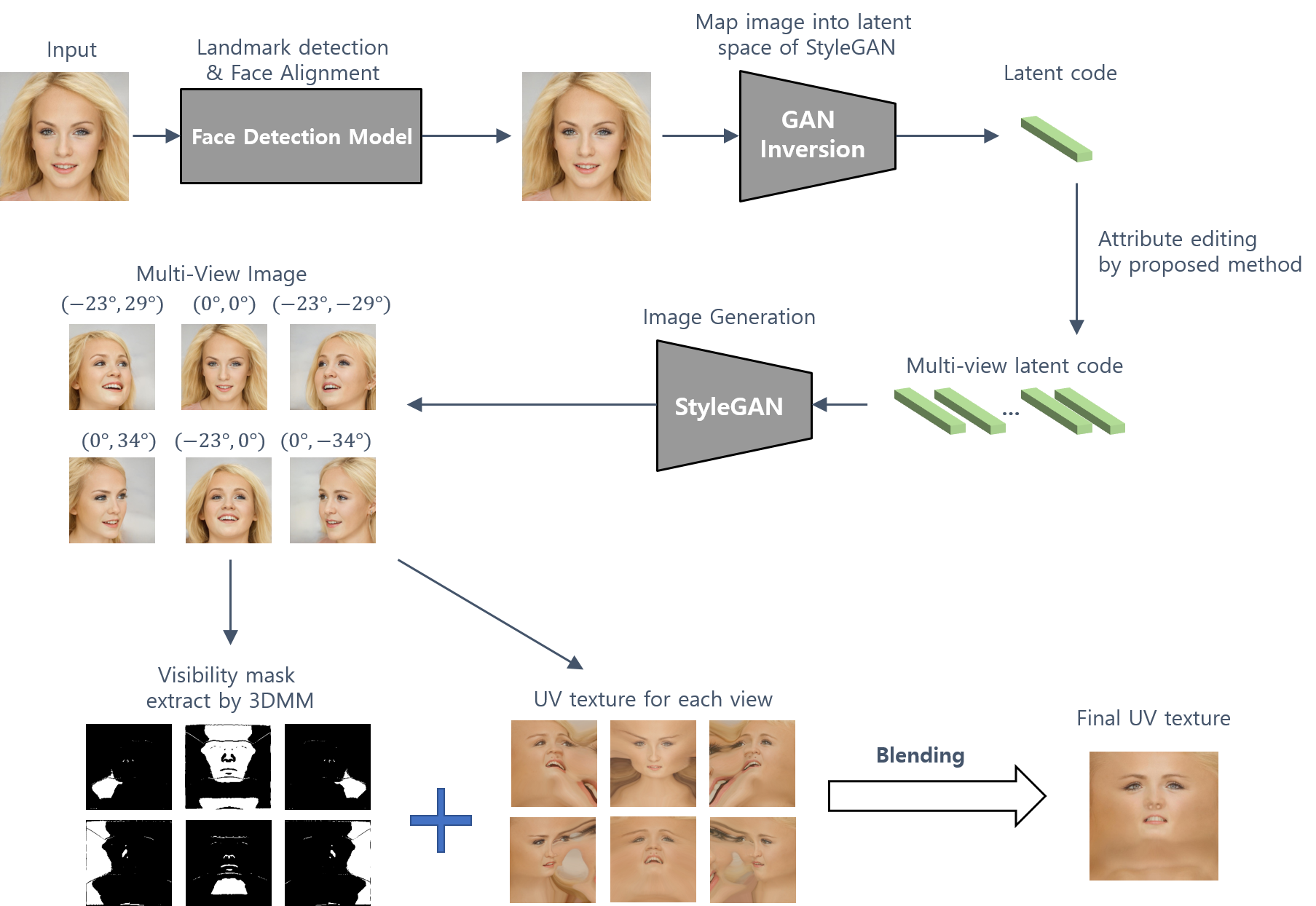}
\end{center}
\setlength{\abovecaptionskip}{0pt}
   \caption{Schematic diagram of UV map generation process.}
\label{uv_process}
\end{figure}

\subsection{Architecture Details}
All pretrained models are obtained from the official website. The details of ‘Map and edit’ is shown in Fig. \ref{mlp}, both the $M_f$ and $M_i$ are implemented using an 5-layer MLP. The $M_f$ and the $M_i$ are encoder and decoder respectively. 

\section{UV Map Generation Process}
See Fig. \ref{uv_process}, UV map generation process is composed of two steps as follow:

(1) Multi-view face image generation: Given a input image $I$, we utilize the GAN inversion model $F$ \cite{tov2021designing} to obtain $w=F(I)$. Then, we generate face images $\{I_i\}_0^5=\{G(w_i)\}_0^5$ with specified angles by the ‘Map and edit’.

(2) Synthesize UV texture with multi-view images: we extract the visibility mask $\mathbb{M}=\{m_i\}_0^5 \in \mathbb{R}^{1\times1024\times1024}$ and UV map $\textbf{U}=\{U_i\}_0^5 \in \mathbb{R}^{3\times1024\times1024}$ similar to \cite{gecer2021ostec}. Here, the visibility mask is defined as maintaining optimal visibility of each part of the face in a near-vertical view. The final UV map $U_f=\textbf{U}\odot\textbf{M}$ generated by blending the visibility mask and UV map.

\section{Additional Experimental Results}

\subsection{Simultaneous Parameter Control}

\begin{figure}[!t]
\begin{center}
\centering\includegraphics[scale=0.4]{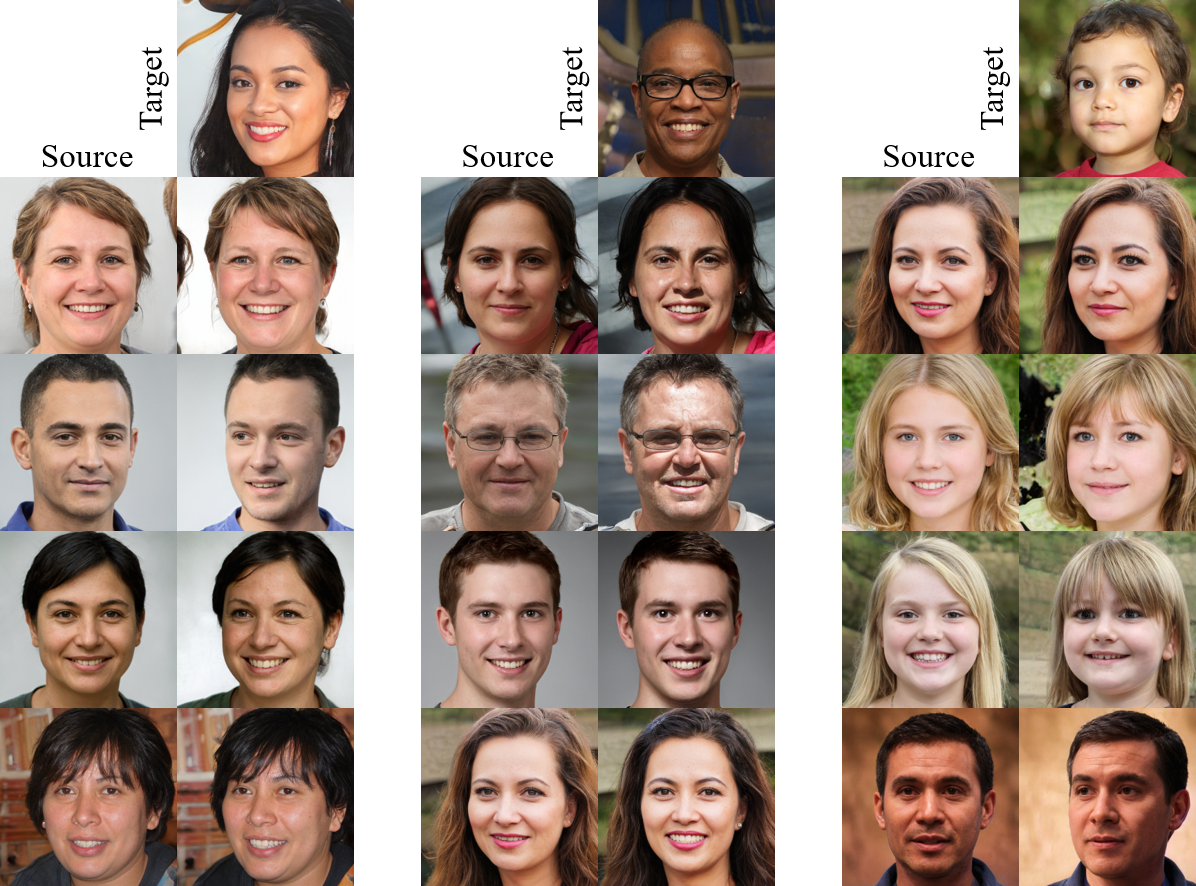}
\end{center}
\setlength{\abovecaptionskip}{0pt}
   \caption{The result of blending the pose, lighting and expression obtained from the target image to the source image.}
\label{supp6}
\end{figure}

Fig. \ref{supp6} shows mixing results where pose, expression and illumination parameters are transferred from the source to target images.

\begin{figure}[!t]
\begin{center}
\centering\includegraphics[scale=0.44]{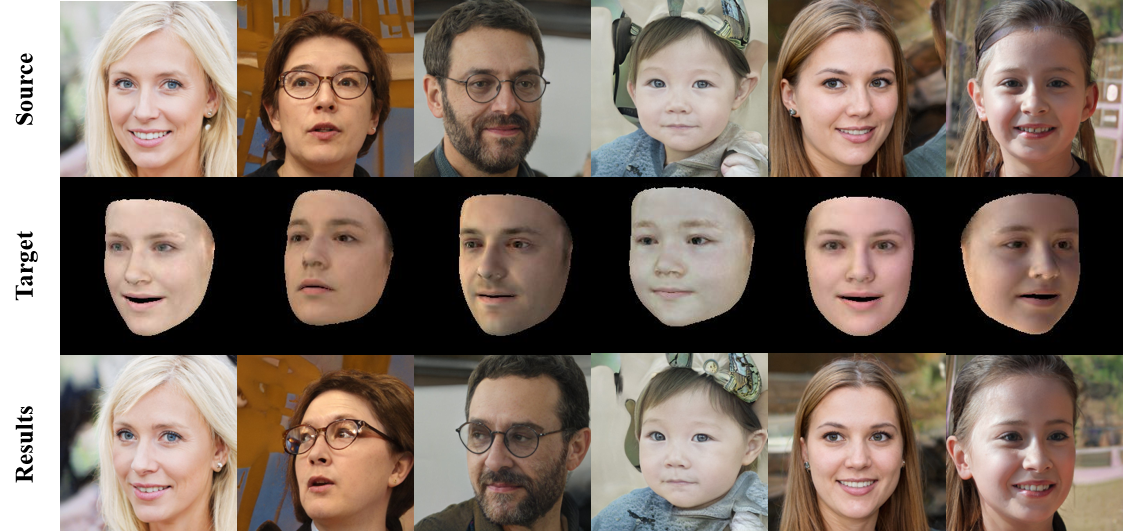}
\end{center}
\setlength{\abovecaptionskip}{0pt}
   \caption{Results of 3DMM guided pose editing.}
\label{supp2}
\end{figure}

\begin{figure}[!t]
\begin{center}
\centering\includegraphics[scale=0.44]{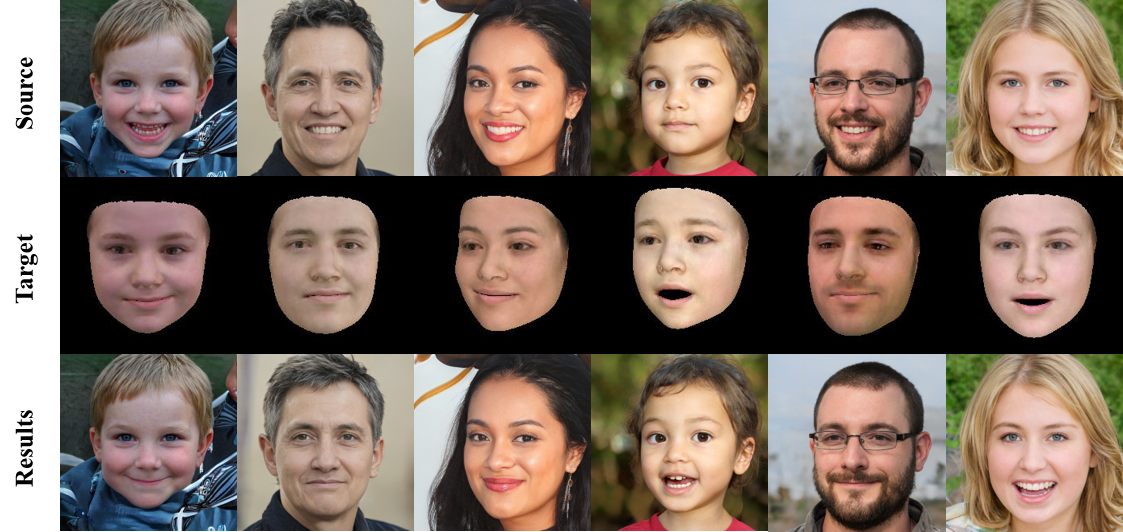}
\end{center}
\setlength{\abovecaptionskip}{0pt}
   \caption{Results of 3DMM guided expression editing.}
\label{supp3}
\end{figure}

\begin{figure}[!t]
\begin{center}
\centering\includegraphics[scale=0.44]{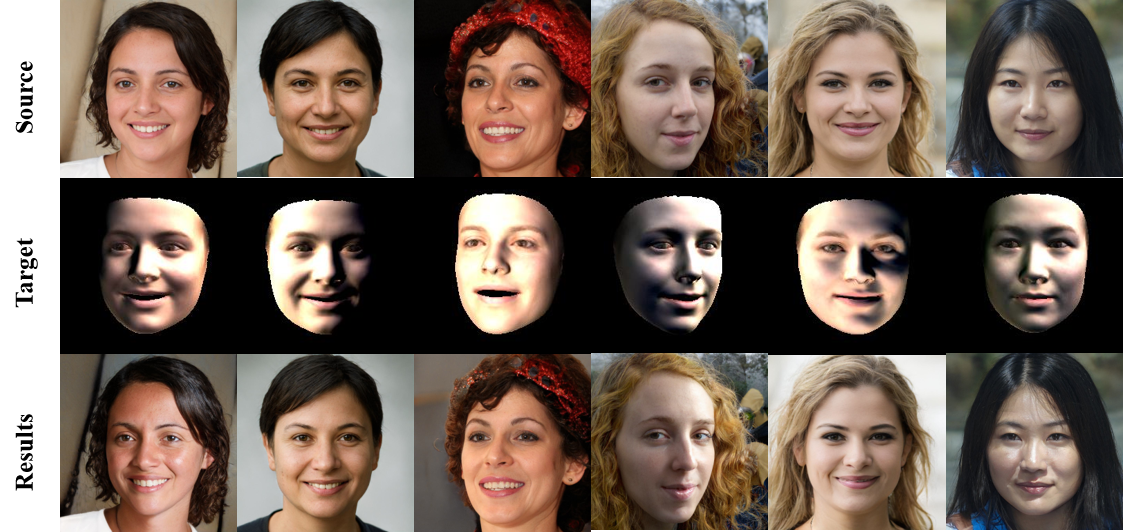}
\end{center}
\setlength{\abovecaptionskip}{0pt}
   \caption{Results of 3DMM guided lighting editing.}
\label{supp4}
\end{figure}

\subsection{3DMM Guided Image Manipulation}
Our image editing is based on explicit control of 3DMM parameters. As described in $Method$ section, we guide 2D generator by editing the $P_e$ obtained from $P$. We show results on target 3DMM and edited 2D images. As shown in Fig. \ref{supp2}, we show results on \textbf{pose} editing. As shown in Fig. \ref{supp3}, we show results on \textbf{expression} editing. As shown in Fig. \ref{supp3}, we show results on \textbf{lighting} editing.

\subsection{Latent Space to 3DMM Results}
\begin{figure*}[!t]
\begin{center}
\centering\includegraphics[width=\textwidth]{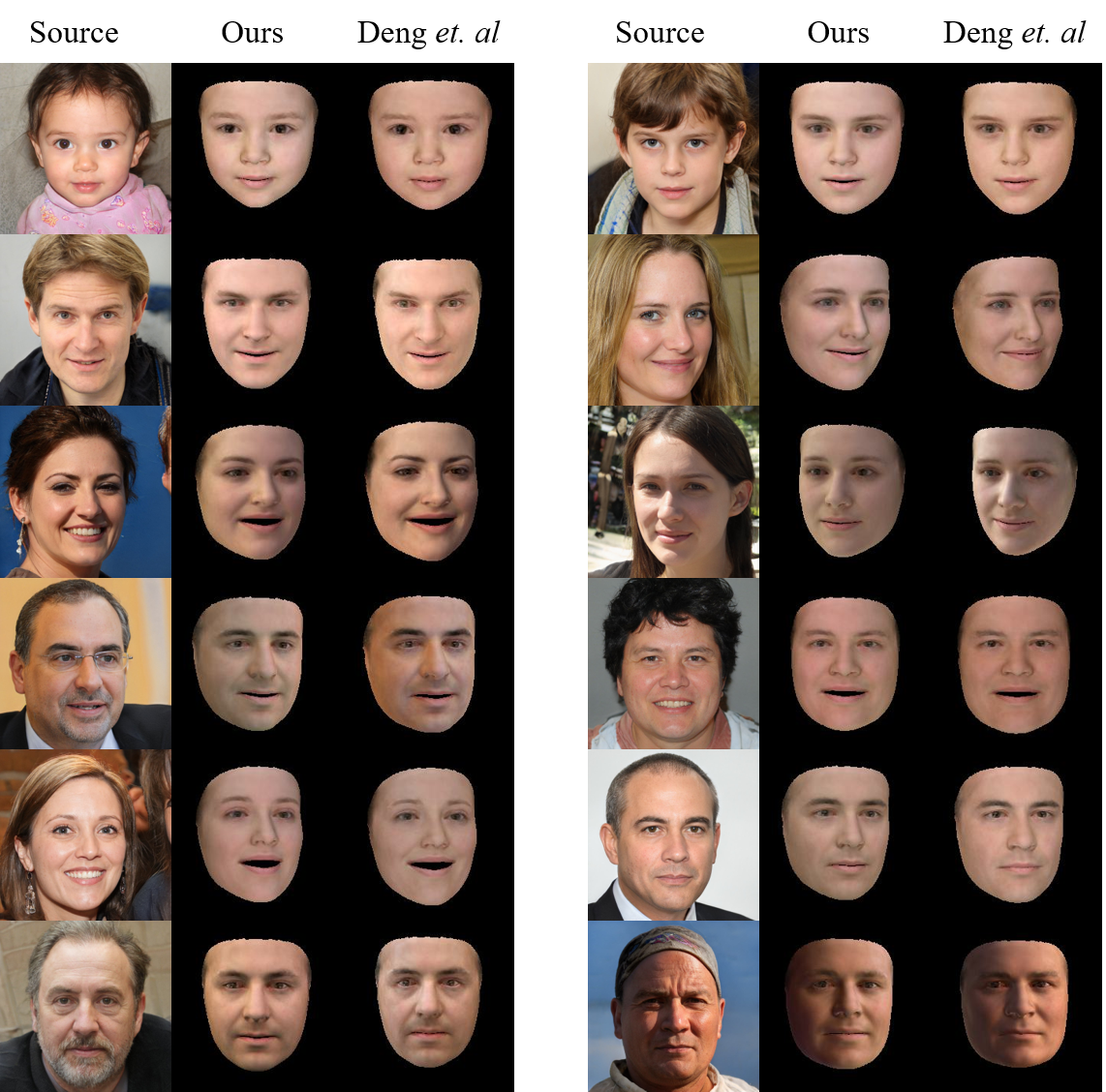}
\end{center}
\setlength{\abovecaptionskip}{0pt}
   \caption{Qualitative comparison of 3DMM reconstruction results.}
\label{supp1}
\end{figure*}

As shown in Fig. \ref{supp1}, we show comparative results for 3DMM reconstructions. The results of \cite{deng2019accurate} are extracted from source images $I_w$. Ours are mapped from latent code $P_w=M_f(w)$ directly. It can be seen from the results that although there will be some differences in texture and lighting, the consistency in pose is very high.

\subsection{CelebA-HQ Face UV Map Completion}

\begin{figure*}[!t]
\begin{center}
\centering\includegraphics[width=\textwidth]{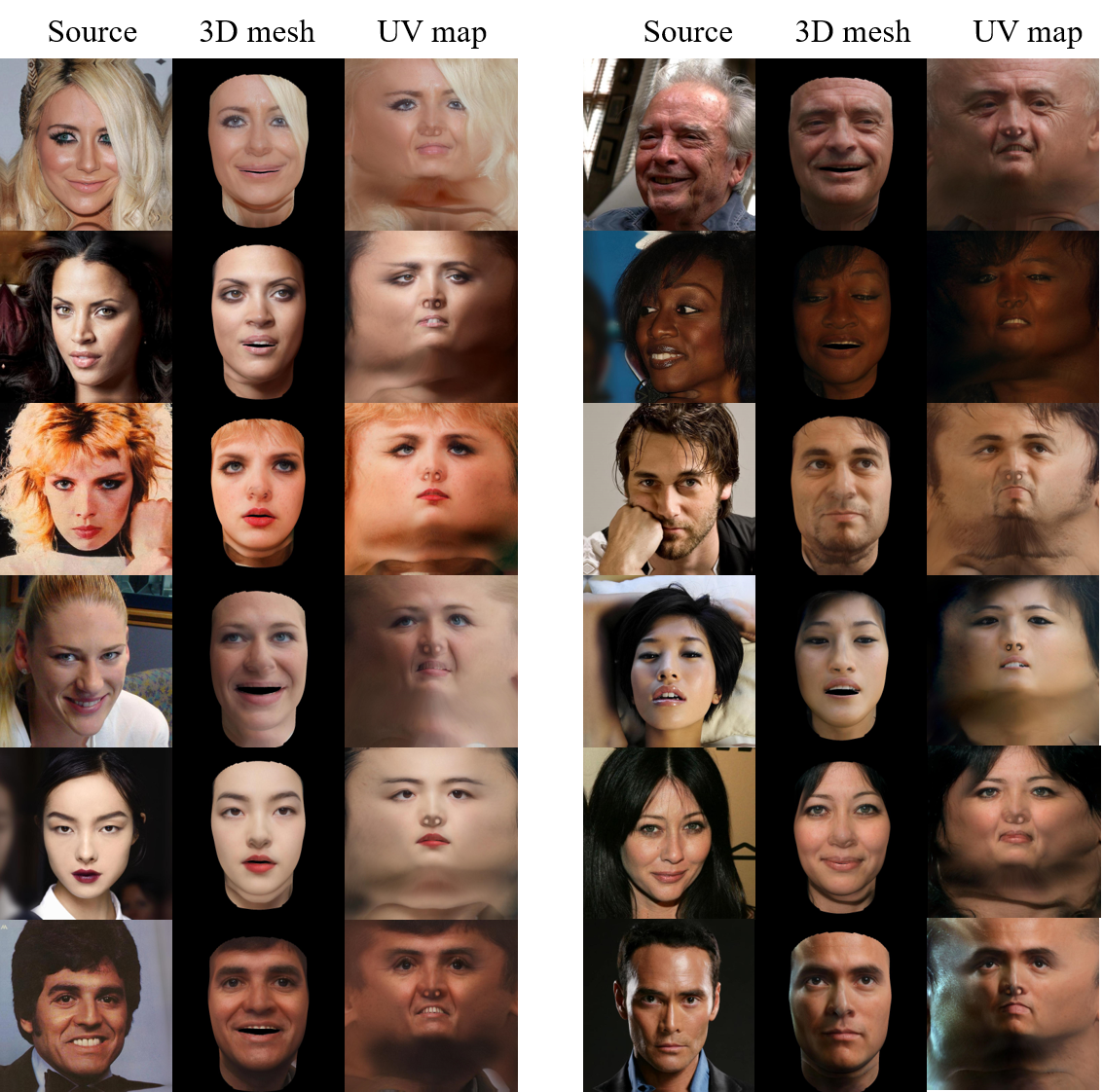}
\end{center}
\setlength{\abovecaptionskip}{0pt}
   \caption{UV map completion results on CelebA-HQ.}
\label{supp5}
\end{figure*}

As shown in Fig. \ref{supp5}, we show more face UV map generation results on CelebA-HQ dataset \cite{karras2017progressive}. It can be seen from the results that our model still has great performance when the input images has different degrees of unseen regions.

\subsection{Real-World Face UV Map Completion}

\begin{figure*}[!t]
\begin{center}
\centering\includegraphics[width=\textwidth]{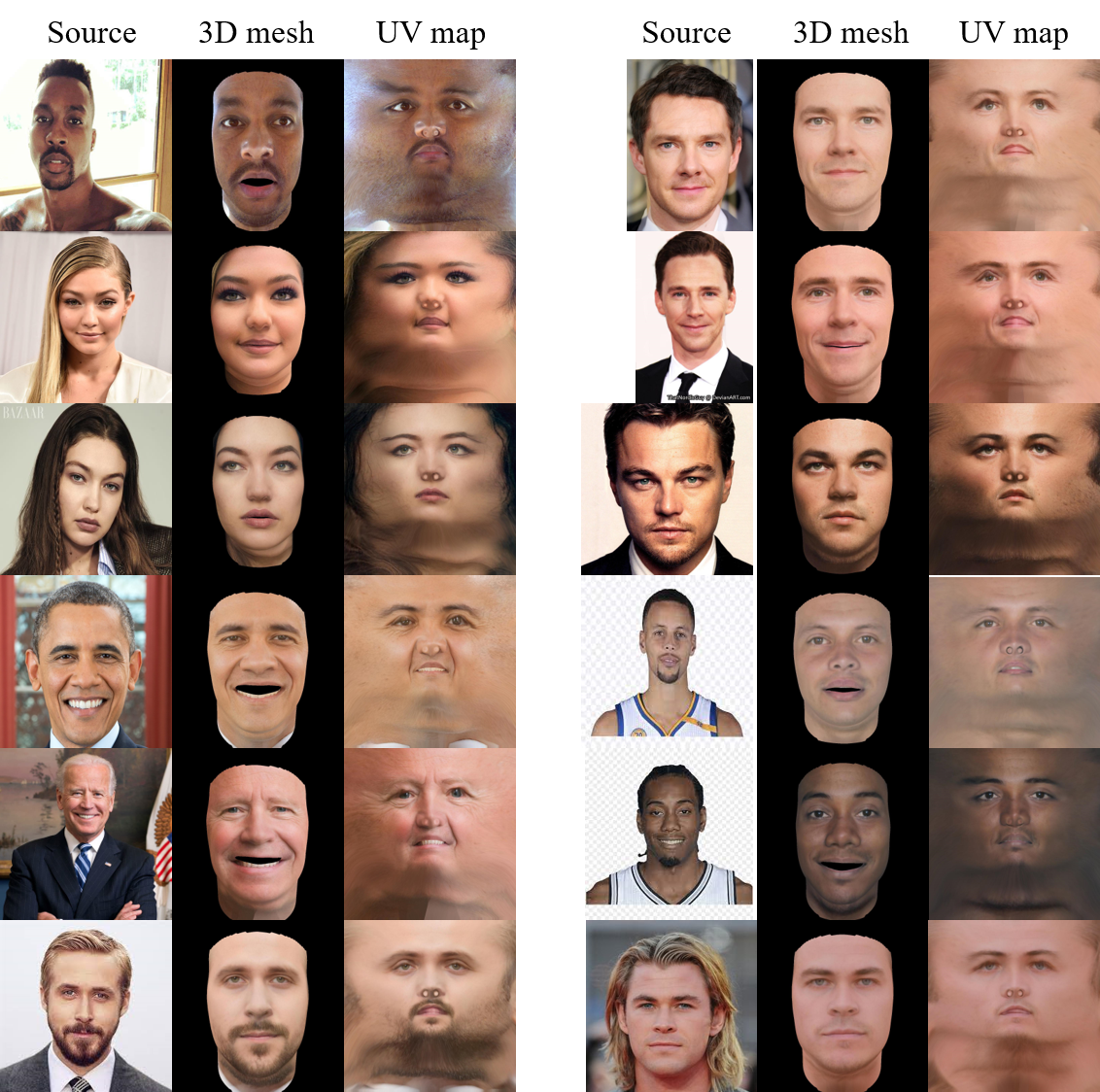}
\end{center}
\setlength{\abovecaptionskip}{0pt}
   \caption{UV map completion results on real-world images.}
\label{supp7}
\end{figure*}

As shown in Fig. \ref{supp7}, we show more face UV map generation results on real-world images. In the UV map generation process for real-world images, we employ HyperStyle\cite{alaluf2021hyperstyle} as the model for GAN inversion.

\subsection{Video Sample}
We also provide a video showing our edits on pose, expression, lighting, and UV map results. Our results show that by explicitly editing the parameters of the 3DMM, our 2D images can be edited smoothly.

\section{Limitations}

Our approach is limited by the image generation capability of StyleGAN2 and the quality of the 3D face reconstruction network. Secondly, since the 3DMM has no parameters about glasses, when we edit the pose of the image, there is a small probability that the glasses may appear or disappear. Finally, our model does not impose constraints on factors such as background hairstyles, therefore, these parts are not controllable and may change when editing parameters.

\bibliography{aaai22}